\documentclass[10pt, conference]{IEEEtran}
\newcommand{\matr}[1]{\mathbf{#1}} 
\usepackage{amsmath,amsfonts,bm}

\def\eqref#1{equation~\ref{#1}}
\def\1{\bm{1}}

\DeclareMathAlphabet{\mathsfit}{\encodingdefault}{\sfdefault}{m}{sl}
\SetMathAlphabet{\mathsfit}{bold}{\encodingdefault}{\sfdefault}{bx}{n}

\usepackage{graphicx}

\usepackage{amsmath}
\usepackage{amsfonts}     

\usepackage{amsthm}
\usepackage{nicefrac,placeins,flushend}

\usepackage{amssymb}

\usepackage{hhline}
\usepackage{array}

\usepackage{multirow}
\usepackage{tabularx}

\usepackage{diagbox}

\usepackage{xcolor}
\usepackage{adjustbox} 
\usepackage{booktabs}     
\usepackage{subfig}
\usepackage{float}

\ifCLASSINFOpdf
\else
\fi
\begin{document}

% Page Style
%\pagestyle{plain}

%\title{Bare Demo of IEEEtran.cls\\ for IEEE Conferences}

\title{EPE-NAS: Efficient Performance Estimation Without Training for Neural Architecture Search} %EPEWOT-NAS 
%NAS-EWOT Efficient Neural Architecture Search Without Training}
%Neural Architecture Search: Efficient Performance Estimation Without Training

\newcommand\blfootnote[1]{%
  \begingroup
  \renewcommand\thefootnote{}\footnote{#1}%
  \addtocounter{footnote}{-1}%
  \endgroup
}

% DOUBLEBLIND
\author{\IEEEauthorblockN{ Vasco Lopes\IEEEauthorrefmark{1}, Saeid Alirezazadeh \IEEEauthorrefmark{2,3},   Lu{\'{\i}}s A. Alexandre\IEEEauthorrefmark{1}}
\IEEEauthorblockA{\IEEEauthorrefmark{1}NOVA LINCS, Universidade da Beira Interior}
\IEEEauthorblockA{\IEEEauthorrefmark{2}C4-Cloud Computing Competence Center, Universidade da Beira Interior}
%Rua Marqu\^{e}s d'\'{A}vila e Bolama, 6201-001, Covilh\~{a}, Portugal\\
\{vasco.lopes, luis.alexandre\}@ubi.pt \qquad \{saeid.alirezazadeh\}@gmail.com
}


\newcommand{\ver}[1]{{\color{red}!!! #1 !!!}}  % \ver{}

\maketitle


\begin{abstract}
Neural Architecture Search (NAS) has shown excellent results in designing architectures for computer vision problems. NAS alleviates the need for human-defined settings by automating architecture design and engineering. However, NAS methods tend to be slow, as they require large amounts of GPU computation. This bottleneck is mainly due to the performance estimation strategy, which requires the evaluation of the generated architectures, mainly by training them, to update the sampler method. In this paper, we propose EPE-NAS, an efficient performance estimation strategy, that mitigates the problem of evaluating networks, by scoring untrained networks and creating a correlation with their trained performance. We perform this process by looking at intra and inter-class correlations of an untrained network. We show that EPE-NAS can produce a robust correlation and that by incorporating it into a simple random sampling strategy, we are able to search for competitive networks, without requiring any training, in a matter of seconds using a single GPU. Moreover, EPE-NAS is agnostic to the search method, since it focuses on the evaluation of untrained networks, making it easy to integrate into almost any NAS method.
\end{abstract}


% \fi
\IEEEpeerreviewmaketitle

%%%%%%

\section{Introduction}
In the past years, deep learning algorithms have been extensively researched, and efficiently applied to various tasks with excellent results \cite{deng2014deep, goodfellow2016deep}, especially those related to computer vision \cite{voulodimos2018deep}. The great success in computer vision tasks is mainly attributed to the advent of Convolutional Neural Networks (CNNs) \cite{khan2020survey}, given their robust feature extraction capability and transferability between different problems. Different CNNs architectures have gradually been proposed, incrementally showing that CNNs can be improved, by revising the architecture itself, adding additional components such as residual connections, reducing the number of parameters, the size or inference time \cite{DBLP:conf/nips/KrizhevskySH12, DBLP:journals/corr/SimonyanZ14a, DBLP:conf/cvpr/SzegedyLJSRAEVR15, DBLP:conf/cvpr/HeZRS16, DBLP:conf/cvpr/HuangLMW17, DBLP:conf/icml/TanL19}. However, designing efficient architectures is extremely time-consuming. It requires expert knowledge and trial and error. Deep neural networks can have many design choices, such as layers, their combination and sequence, parameters associated with the layers, architecture, and the training procedure as well as optimization rules. Therefore, an automated way to conduct neural architectures' design came as a natural process \cite{hutter2019automated}.

\begin{figure}[!t]
    \centering
    \includegraphics[width=1\columnwidth]{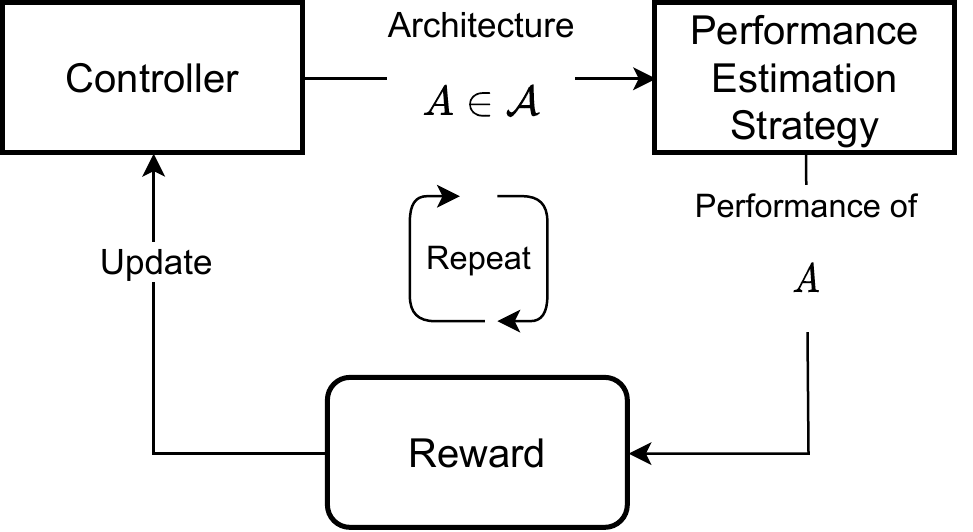}
    \caption{General Neural Architecture Search flow. A controller generates an architecture $A$ from the search space of possible operations and architectures, $\mathcal{A}$, which is then evaluated, and its performance is used as reward to update the controller. Our method acts in the performance estimation block, by obtaining estimates without training. \label{fig:nasdiagram}}
\end{figure}

% , semantic segmentation \cite{liu2019auto}, object detection \cite{chen2019detnas}, image generation \cite{gao2020adversarialnas}, among others. 
Neural Architecture Search (NAS) aims to automate architecture engineering and design, by autonomously designing high performance architectures for a given problem \cite{elsken2019neural}. NAS methods for computer vision problems have been successfully applied to various tasks, such as image classification, semantic segmentation, object detection, and others \cite{elsken2019neural, DBLP:journals/corr/abs-1905-01392}. Since the incipience proposal \cite{DBLP:conf/iclr/ZophL17}, NAS methods broadly focused on designing architectures using a similar flow. A controller, using a specified search strategy, being the most common Reinforcement Learning or Evolutionary Strategies, samples an architecture $A$ from the space of possible architectures $\mathcal{A}$, which is defined by the search space, that comprises the possible operations (e.g., convolution, pooling) and the architecture type. The generated architecture is evaluated, and the result is given as a reward to the controller to update its parameters. This process is repeated thousands of times, whereby the controller learns to sample better architectures over time. A visualization of this process can be seen in Fig. \ref{fig:nasdiagram}.

Although NAS methods have shown excellent results, the computational cost of most methods is extremely high, which in some cases can be in the order of months of GPU computation \cite{DBLP:conf/iclr/ZophL17, liu2018progressive, Zoph_2018}. This is mainly associated with the performance estimation strategy, which evaluates the generated architectures based on regular training, either from scratch until convergence or partial training \cite{runge2018learning, real2019aging}. Recent approaches, attempt to smooth the training process, by sharing parameters \cite{pham2018efficient}, applying mutations to already trained networks \cite{DBLP:conf/iclr/ElskenMH19}, or by using one-shot NAS, where the controller generates architectures and corresponding weights \cite{DBLP:conf/iclr/LiuSY19, DBLP:conf/iclr/CaiZH19, DBLP:conf/iclr/ZelaESMBH20}. However, some NAS proposals have shown to be overfitting the search space and not allowing exploration due to the introducing of design bias \cite{Yang2020NAS}.

To mitigate the aforementioned problems, in this paper, we propose EPE-NAS, a performance estimation strategy that scores generated networks at initialization stage, without requiring any training. By evaluating how the gradients of the network behave with respect to the input, it is possible to score \textbf{untrained} networks, eliminating the need to train generated architectures to update parameters. The proposed method is extremely fast, allowing the analysis of thousands of networks in seconds. We show that this method can be used to guide the search over the search space due to its fast inference of a network trained accuracy from its untrained state. The proposed method can be easily integrated into almost any NAS method, by entirely replacing the performance estimation strategy, or complementing it, by creating a multi estimation strategy. We show this by incorporating the proposed method into a random search strategy, achieving competitive results in seconds. The code for the proposed method is also available\footnote{Code publicly available on GitHub: www.github.com/VascoLopes/EPENAS}.

The main contributions of this paper can be summarized as follows:
\begin{itemize}
    \item We propose a novel performance estimation strategy that can evaluate the trained performance of an  untrained network, which can be easily integrated into almost any NAS method.
    \item We analyze the impact of the proposed method when coupled with random search, showing that it can achieve competitive results in a few seconds.
    \item We compare the proposed method with different NAS methods, as well as with a surrogate performance estimator in NAS-Bench-201.
    \item We show that the proposed method allows the search space to be quickly analyzed without the need to train networks, allowing bad candidates to be weed out, by analyzing the relationship between the score and the network performance when trained.
\end{itemize}

%%%%ver
The remainder of this work is organized as follows. Section~\ref{related_work}, contextualizes the related work. Section~\ref{proposed_method}, describes the proposed method in detail. In Section~\ref{experiments}, we present the experiments performed, the datasets and benchmark used, the results and discussion. Finally, in Section~\ref{conclusions}, a conclusion is drawn.

%%%%%%%%%%%%%%%%%%%%%%%%%%%%%%%%%%%%%%%%%%%%%%%%%%%%%

\section{Related Work}
\label{related_work}
Generally, NAS methods attempt to automatically design optimal CNNs using a sample-evaluate-update scheme, where a controller generates an architecture and is updated using the generated architecture performance. The problem with this is that evaluating the generated architectures is very costly. Zoph and Le \cite{DBLP:conf/iclr/ZophL17} initially formulated NAS as a reinforcement learning problem, where a controller was trained over-time to sample more efficient architectures. The problem was that this method required more than 60 years of GPU computation, as it trained all generated architectures to convergence. As follow-up work, the authors tackle this problem by performing a cell-based search in a search space with 13 operations \cite{Zoph_2018}. By focusing on designing two types of cells: normal cells (perform convolutional operations) and reduction cells (reduce input size), the authors could reduce the GPU computation to 2000 days. More, they found that cell-based architectures searched in CIFAR-10 can be transferred to ImageNet by stacking more cells. In this method, more than 20000 networks were trained and evaluated.

Similar to \cite{DBLP:conf/iclr/ZophL17}, in \cite{DBLP:conf/iclr/BakerGNR17} the authors present MetaQNN, a method based on reinforcement learning and Q-learning, where the learning agent was trained to sequentially sample CNN layers. Using a similar reinforcement learning approach, BlockQNN \cite{zhong2018practical} focuses on sampling blocks of operations used to form entire networks. However, BlockQNN still required 96 GPU days of computation. ENAS \cite{pham2018efficient}, used a controller, trained with policy gradient, to discover architectures by searching for an optimal subgraph within a large computational graph. By constructing a sizeable computational graph, where each subgraph represents a network, ENAS forced all generated architectures to share their parameters. In this way, efficient search was enabled in less than one GPU computational day. The authors of \cite{DBLP:conf/iclr/LiuSY19} proposed DARTS, a gradient-based method, that by performing continuous relaxation of the search space to be continuous, it optimized architectures using gradient descent. The authors propose a bilevel gradient optimization, which jointly learns the architecture and the weights in a few GPU days. This paper served as foundation for many other one-shot methods. In \cite{Zela2020Understanding}, the authors improve DARTS generated architecture performances by introducing regularization mechanisms. Also using a differentiable approach, GDAS \cite{dong2019searching} is a method that makes the search procedure differentiable, so that sub-graphs can be sampled from the directed acyclic graph representing the search space, which can be trained end-to-end to sample efficient networks. GDAS' controller is optimized based on the validation loss of the trained sampled architecture. SETN \cite{DBLP:conf/iccv/Dong019a}, also uses a differentiable approach, but uses an evaluator trained to indicate the probability of each architecture to have a low validation loss, which allows selective sampling of networks. REA \cite{DBLP:conf/aaai/RealAHL19}, instead of reinforcement learning or gradient-based methods, focuses on using evolutionary tournament selection algorithms with an age property that favors younger architectures.

To mitigate the bottleneck of the performance estimation strategy, surrogate methods have also been proposed to extrapolate the learning curve with a partial train \cite{domhan2015speeding,DBLP:conf/iclr/BakerGRN18}, or by learning a HyperNet that generates weights based on the architecture \cite{DBLP:conf/iclr/BrockLRW18}. In \cite{DBLP:conf/icml/FalknerKH18}, the authors propose BOBH, that focuses on hyperparameter optimization, which includes architecture design, by combining Bayesian optimization and bandit-based methods. However, it still required 33 GPU days to design an optimal network in CIFAR-10.

Our work differentiates from the aforementioned, being closer to NAS-WOT \cite{mellor2020neural}, as our focus is to evaluate a generated network, without requiring any training, neither for the performance estimation strategy, nor for the generated networks. Thus creating a score that correlates an \textbf{untrained} network to its performance once trained, in efficient time.

%Many NAS approaches suffer from time bottlenecks mainly due to the performance estimation strategy. Initially, methods some greatly improve this constraint by using Surrogate models \cite{lu2020nsganetv2}, allowing the proposed methods to estimate the performace of generated networks without requiring  ,  

%%%%%%%%%%%%%%%%%%%%%%%%%%%%%%%%%%%%%%%%%%%%%%%%%%%%%%%%%%%%%%%%%%%%%%%%%%%%%%%%%%%%%%%
\section{Proposed Method}
\label{proposed_method}
In this paper, we propose EPE-NAS, a novel performance estimation strategy, whose goal is to estimate the performance of generated networks without requiring any training, neither for the generated networks nor for the performance estimator. To do this, we score \textbf{untrained} networks as an indicator of their accuracy when trained.

We base our approach on the idea proposed in \cite{mellor2020neural}, which states that different networks can be compared by evaluating their behavior using local linear operators at different data points. The local linear operators are obtained by multiplying the linear maps at each layer interspersed with the binary rectification units. %To do this, let $w_i$ denote the linear map which maps $x_i \in \mathbb{R}^{D}$ to a final representation $f(x_i)$, where D is the input dimension.
To do this, one can define a linear map, $w_i=f(x_i)$, which maps the input $\mathbf{x}_i  \in \mathbb{R}^{D}$, through the network, $f(\mathbf{x}_i)$, where $\mathbf{x}_i$ represents an image that belongs to a batch $\mathbf{X}$, and $D$ is the input dimension. Then, the linear map can be computed using:
\begin{equation*}
    Jacobian~\mathbf{w}_i = \frac{\partial f(\mathbf{x}_i)}{\partial \matr{x_{i}}}
\end{equation*}

%one can define a linear map, which maps the input, $\mathbf{x}_i  \in \mathbb{R}^{D}$, through the network, $f(\mathbf{x}_i)$, where $\mathbf{x}_i$ represents an image that belongs to a batch $\mathbf{X}$, and $D$ is the input dimension. Then, the linear map can be computed using the Jacobian $\mathbf{w}_i$ = $\frac{\partial f(\mathbf{x}_i)}{\partial \matr{x_{i}}}$.

In order to evaluate how a network behaves with different data points, we calculate the Jacobian matrix $\mathbf{w}_i$ for different data points, $f(\mathbf{x}_i)$, of the batch $\mathbf{X}$, $i \in 1, \cdots, N$:
\begin{equation*}
\matr{J} = 
\begin{pmatrix}
\frac{\partial f(\mathbf{x}_1)}{\partial \matr{x_{1}}} & \frac{\partial f(\mathbf{x}_2)}{\partial \matr{x_{2}}} & \cdots & \frac{\partial f(\mathbf{x}_N)}{\partial \matr{x_{N}}} \\
\end{pmatrix}^{\top}
\label{eq:J}
\end{equation*}

The Jacobian Matrix $\matr{J}$ contains information about the network output with respect to the input for several data points. We then can evaluate how points belonging to the same class correlate with each other, where the goal is to see if an \textbf{untrained} network is capable of modeling complex functions. Explicitly, a flexible network should simultaneously be able to distinguish local linear operators for each data point, but also have similar results for similar data points, which in a supervised approach means that the data points belong to the same class. The perfect scenario would be to have an \textbf{untrained} network with low correlation between different data points, where data points of the same category are closer to each other, which means that the network would easily learn to distinguish the two data points during training. To evaluate this behavior, we evaluate the correlation of $\matr{J}$ values with respect to their class, by computing a covariance matrix for each class present in $\matr{J}$: $\matr{C}_{J_c} = (\matr{J} - \matr{M}_{J_c})(\matr{J}-\matr{M}_{J_c})^\top$, where $\matr{M}_J$ is the matrix with elements: %$(\matr{M}_{J_c})_{i,j} = \frac{1}{N}\sum_{n\in{1,\ldots,N},\\ x_i \in class c}J_{i,n}$
%$\displaystyle (\matr{M}_{J_c})_{i,j} =\mathop{\sum_{n\in{1,\ldots,N}}}_{x_i \in class c} J_{i,n} $
\begin{equation*}
    (\matr{M}_{J_c})_{i,j}=\frac{1}{N}
\sum_{\substack{n\in\{1,\ldots,N\},\\ x_i \in class~c}}
\matr{J}_{i,n},
\end{equation*}

\noindent where $c$ represents the class, $c \in 1, ..., C$, and $C$ is the number of classes present in the batch. Then, it is possible to calculate the correlation matrix per class, $\matr{\Sigma}_{J_c}$, for each covariance matrix $\matr{C}_{J_c}$: $({{\matr{\Sigma}_J}_c})_{i,j} = \frac{(\matr{C}_{J_c})_{i,j}}{\sqrt{(\matr{C}_{J_c})_{i,i} * (\matr{C}_{J_c})_{j,j}}}$, where $(i,j)$ represents the $(i,j)^{th}$ element of the matrices.

Each individual correlation matrix allows the analysis of how the \textbf{untrained} network behaves for each class, which may be an indication of the ability of the local linear operators to perceive differences between classes.

To allow comparison between the different individual correlation matrices, as they may have different sizes due to the number of data points per class, they are individually evaluated:
\begin{equation*}
    \matr{E}_c =  
    \begin{cases}
        \sum_{i=0}^{N}\sum_{j=0}^{N} log(|({{\matr{\Sigma}_J}_c})_{i,j}|+k), & \text{if }C \leq 100\\\\
        \frac{\sum_{i=0}^{N}\sum_{j=0}^{N} log(|({{\matr{\Sigma}_J}_c})_{i,j})|+k}{||{{\matr{\Sigma}_J}_c}||}, & \text{otherwise}
    \end{cases}
\end{equation*}

\noindent where $k$ is a small-constant with the value of $1\times10^{-5}$, and $C$ is the number of classes in batch $\mathbf{X}$. To avoid confusion with absolute value operation, we denote $||X||$ as the number of elements of the set X.
The normalization based on the size of the correlation matrix is due to the fact that for a constant batch size, as the number of classes increases, the size of the individual correlation matrices becomes smaller, a correlation matrix with a larger size would obtain a larger value.

Then, a network is scored based on the individual evaluations of the correlation matrices by:

\begin{equation*}
    S =  
    \begin{cases}
        \sum_{t=0}^{C} |\matr{E}_t|, & \text{if }C \leq 100\\\\
        \frac{\sum_{i=0}^{C}\sum_{j=i}^{C} |\matr{E}_i - \matr{E}_j|}  {||\matr{E}||} , & \text{otherwise}
    \end{cases}
\end{equation*}

\noindent where E is the vector containing all the correlation matrices' scores. Depending on the number of classes present in the batch, the final score is either a sum of the individual correlation matrices' scores or a normalized pair-wise difference. Normalization serves to mitigate the class difference when evaluating networks in datasets with a high number of classes and noise.

% \begin{figure*}[t]
%   \centering
%   \subfloat[CIFAR-10]{\includegraphics[width=0.28\textwidth]{images/cells/cell_cifar10.pdf}\label{fig:cell_cifar10}}
%   \subfloat[CIFAR-100]{\includegraphics[width=0.28\textwidth]{images/cells/cell_cifar100.pdf}\label{fig:cell_cifar100}}
%   \subfloat[ImageNet16-120]{\includegraphics[width=0.28\textwidth]{images/cells/cell_imagenet.pdf}\label{fig:cell_imnet}}
%   \subfloat{\includegraphics[width=0.15\textwidth]{images/cells/legend.pdf}\label{fig:f2}}
%   \caption{My flowers.}
% \end{figure*}

\hfill
%%%%%%%%%%%%%%%%%%%%%%%%%%%%%%%%%%%%%%%%%%%%
\section{Experiments}
\label{experiments}
We evaluate the effectiveness of EPE-NAS on three datasets: CIFAR-10, CIFAR-100 and ImageNet16-120 from NAS-Bench-201, using a batch size of 256. As the proposed method is a performance estimation strategy that does not require any training, we also evaluate EPE-NAS by combining it with a random search strategy \cite{li2020random}, where a candidate network from the search space is randomly proposed and scored using the proposed performance estimation method, instead of training the network. This evaluation is done for different sample sizes of $N$ networks.

The setup for all the experiments conducted was a desktop computer, with a single 1080Ti GPU and 32GB of ram. In the following sections, we comment on NAS-Bench-201, and individually detail the experiments, results and provide a discussion.

\begin{table*}[htb]
	\caption{Comparison of several search methods evaluated using the NAS-Bench-201 benchmark. Performance shown in accuracy with mean$\pm$std, on CIFAR-10, CIFAR-100 and ImageNet-16-120. Methods are divided into 4 blocks, depending on their approach: weight sharing, non-weight sharing, training-free approaches (with a direct comparison between the proposed method and NAS-WOT), and a baseline using an SVM as a surrogate estimator. 
		Search times are the mean time required to search for cells in CIFAR-10, using a single 1080Ti GPU. Search time includes the time taken to train networks as part of the process where applicable. The performances of the training-free approaches are given for different sample size~\texttt{N}. For each sample size, we also report the optimal network. Table adapted from \cite{mellor2020neural}, with reported results for non-weight and weight sharing methods from~\cite{Dong2020NAS-Bench-201}.}
	\vspace{2mm}
	\label{table:benchmarking}
	\footnotesize
	
	\setlength{\tabcolsep}{3pt}
	\setlength{\arrayrulewidth}{2pt}
	
	\renewcommand{\arraystretch}{1.3}
	\resizebox{\textwidth}{!}{%
		\begin{tabular}{@{}lr@{\hskip 0.15in}llcllcll@{}} \hline 
			\multirow{2}{*}{Method} & \multirow{2}{*}{\shortstack{Search \\Time (s)}}  & \multicolumn{2}{c}{CIFAR-10} & \phantom{ab} & \multicolumn{2}{c}{CIFAR-100} & \phantom{ab} & \multicolumn{2}{c}{ImageNet-16-120} \\
			\cmidrule{3-4} \cmidrule{6-7} \cmidrule{9-10}
			& & \multicolumn{1}{c}{validation} & \multicolumn{1}{c}{test} && \multicolumn{1}{c}{validation} & \multicolumn{1}{c}{test} && \multicolumn{1}{c}{validation} & \multicolumn{1}{c}{test} \\
			\midrule
			\multicolumn{10}{c}{\textbf{Non-weight sharing}}\\
			REA       &  12000 & 91.19$\pm$0.31 & 93.92$\pm$0.30 && 71.81$\pm$1.12 & 71.84$\pm$0.99 && 45.15$\pm$0.89 & 45.54$\pm$1.03 \\
			RS        &  12000 & 90.93$\pm$0.36 & 93.70$\pm$0.36 && 70.93$\pm$1.09 & 71.04$\pm$1.07 && 44.45$\pm$1.10 & 44.57$\pm$1.25 \\
			REINFORCE &  12000 & 91.09$\pm$0.37 & 93.85$\pm$0.37 && 71.61$\pm$1.12 & 71.71$\pm$1.09 && 45.05$\pm$1.02 & 45.24$\pm$1.18 \\
			BOHB      &  12000 & 90.82$\pm$0.53 & 93.61$\pm$0.52 && 70.74$\pm$1.29 & 70.85$\pm$1.28 && 44.26$\pm$1.36 & 44.42$\pm$1.49 \\
			\midrule \midrule
			
			\multicolumn{10}{c}{\textbf{Weight sharing}}\\
			RSPS        & 7587  & 84.16$\pm$1.69 & 87.66$\pm$1.69 && 59.00$\pm$4.60 & 58.33$\pm$4.34 && 31.56$\pm$3.28 & 31.14$\pm$3.88 \\
			DARTS-V1    & 10890 & 39.77$\pm$0.00 & 54.30$\pm$0.00 && 15.03$\pm$0.00 & 15.61$\pm$0.00 && 16.43$\pm$0.00 & 16.32$\pm$0.00 \\
			DARTS-V2    & 29902 & 39.77$\pm$0.00 & 54.30$\pm$0.00 && 15.03$\pm$0.00 & 15.61$\pm$0.00 && 16.43$\pm$0.00 & 16.32$\pm$0.00 \\
			GDAS        & 28926 & 90.00$\pm$0.21 & 93.51$\pm$0.13 && 71.14$\pm$0.27 & 70.61$\pm$0.26 && 41.70$\pm$1.26 & 41.84$\pm$0.90 \\
			SETN        & 31010 & 82.25$\pm$5.17 & 86.19$\pm$4.63 && 56.86$\pm$7.59 & 56.87$\pm$7.77 && 32.54$\pm$3.63 & 31.90$\pm$4.07 \\
			ENAS        & 13315 & 39.77$\pm$0.00 & 54.30$\pm$0.00 && 15.03$\pm$0.00 & 15.61$\pm$0.00 && 16.43$\pm$0.00 & 16.32$\pm$0.00 \\
			\midrule
			\midrule
			\multicolumn{10}{c}{\textbf{Training-free}}\\
			NAS-WOT (\texttt{N=10}){$\dagger$}  & 3.1 & 89.56$\pm$0.56 & 92.47$\pm$0.04 && 69.36$\pm$1.55 & 69.20$\pm$1.05 && 42.08$\pm$1.61 & \textbf{42.20}$\pm$\textbf{1.37} \\
			\textbf{Ours} (\texttt{N=10})  & \textbf{2.3}   & 89.90$\pm$0.21 & \textbf{92.63}$\pm$\textbf{0.32}  && 69.78$\pm$2.44  & \textbf{70.10}$\pm$\textbf{1.71} && 41.73$\pm$3.60       & 41.92$\pm$4.25 \\[0.1cm]
			
			NAS-WOT (\texttt{N=100}){$\dagger$} & 25.7 & 89.91$\pm$0.80 & 91.41$\pm$2.24  && 67.13$\pm$4.03 & 67.18$\pm$4.14 && 41.39$\pm$1.13 & \textbf{41.42}$\pm$\textbf{1.53}  \\
			\textbf{Ours} (\texttt{N=100}) & \textbf{20.5}  & 88.74$\pm$3.16 & \textbf{91.59}$\pm$\textbf{0.87}  && 67.28$\pm$3.68 & \textbf{67.19}$\pm$\textbf{3.82}  && 38.66$\pm$4.75       & 38.80$\pm$5.41  \\[0.1cm]
			
			NAS-WOT (\texttt{N=500}){$\dagger$} & 126.8 & 88.73$\pm$0.81 & 91.71$\pm$1.37 && 67.62$\pm$1.61 & 67.54$\pm$2.23 && 39.37$\pm$3.01 & 39.84$\pm$3.68 \\ 
			\textbf{Ours} (\texttt{N=500}) & \textbf{105.8} & 88.17$\pm$1.35 & \textbf{92.27}$\pm$\textbf{1.75} && 69.23$\pm$0.62 & \textbf{69.33}$\pm$\textbf{0.66} && 41.93$\pm$3.19       & \textbf{42.05}$\pm$\textbf{3.09} \\ [0.1cm]
			
			NAS-WOT (\texttt{N=1000}){$\dagger$}& 252.6 & 89.60$\pm$0.90 & 91.20$\pm$2.04 && 68.57$\pm$0.41 & 68.95$\pm$0.72 && 38.01$\pm$1.66 & 38.08$\pm$1.58 \\
			\textbf{Ours} (\texttt{N=1000})& \textbf{206.2} & 87.87$\pm$0.85 & \textbf{91.31}$\pm$\textbf{1.69} && 69.44$\pm$0.83 & \textbf{69.58}$\pm$\textbf{0.83} && 41.86$\pm$2.33       & \textbf{41.84}$\pm$\textbf{2.06} \\
			
			%\textbf{Ours} (\texttt{N=10})  & 2.27   & 89.90$\pm$0.21 & 92.63$\pm$0.32  && 69.78$\pm$2.44  & 70.10$\pm$1.71 && 41.73$\pm$3.60       & 41.92$\pm$4.25 \\
			%\textbf{Ours} (\texttt{N=100}) & 20.47  & 88.74$\pm$3.16 & 91.59$\pm$0.87  && 67.28$\pm$3.68 & 67.19$\pm$3.82  && 38.66$\pm$4.75       & 38.80$\pm$5.41  \\
			%\textbf{Ours} (\texttt{N=500}) & 105.84 & 88.17$\pm$1.35 & 92.27$\pm$1.75 && 69.23$\pm$0.62 & 69.33$\pm$0.66 && 41.93$\pm$3.19       & 42.05$\pm$3.09 \\ 
			%\textbf{Ours} (\texttt{N=1000})& 206.23 & 87.87$\pm$0.85 & 91.31$\pm$1.69 && 69.44$\pm$0.83 & 69.58$\pm$0.83 && 41.86$\pm$2.33       & 41.84$\pm$2.06 \\
			\midrule
			%Random & N/A & 83.20$\pm$13.28 & 86.61$\pm$13.46 && 60.70$\pm$12.55 & 60.83$\pm$12.58 && 33.34$\pm$9.39 & 33.13$\pm$9.66 \\ 
			Optimal (\texttt{N}=10)  & N/A & 90.00$\pm$0.95 & 93.41$\pm$0.45  && 70.11$\pm$1.70  & 70.11$\pm$1.70   && 44.67$\pm$1.87       & 44.67$\pm$1.87 \\
			Optimal (\texttt{N}=100) & N/A & 91.12$\pm$0.11 & 94.12$\pm$0.21  && 72.73$\pm$0.78  & 72.73$\pm$0.78   && 46.31$\pm$0.47       & 46.31$\pm$0.47 \\
			Optimal (\texttt{N}=500) & N/A & 91.15$\pm$0.12 & 94.13$\pm$0.22  && 72.83$\pm$0.64  & 72.83$\pm$0.64   && 46.06$\pm$0.66       & 46.06$\pm$0.66 \\
			Optimal (\texttt{N}=1000)& N/A & 91.24$\pm$0.21 & 94.19$\pm$0.15  && 72.92$\pm$0.53  & 72.92$\pm$0.53   && 46.57$\pm$0.59       & 46.57$\pm$0.59 \\ 
			\midrule
			\midrule
			\multicolumn{10}{c}{\textbf{Surrogate Estimator with Training}}\\
			SVM (\texttt{N}=10)   & 359426.3$\ddag$  & 89.74$\pm$1.10 & 92.80$\pm$0.97  && 65.21$\pm$6.48  & 65.46$\pm$6.37   && 37.50$\pm$8.56       & 37.31$\pm$8.66 \\
			SVM (\texttt{N}=100)  & 359449.4$\ddag$ & 87.03$\pm$2.33 & 92.68$\pm$1.47  && 62.82$\pm$5.75  & 63.25$\pm$5.70   && 41.57$\pm$3.55       & 41.73$\pm$3.55 \\
			SVM (\texttt{N}=500)  & 359547.7$\ddag$& 87.37$\pm$2.63 & 93.05$\pm$0.71  && 66.83$\pm$4.34  & 67.36$\pm$4.28   && 41.84$\pm$1.38       & 41.49$\pm$1.39 \\
			SVM (\texttt{N}=1000) & 359666.2$\ddag$& 87.06$\pm$3.14 & 91.24$\pm$2.28  && 68.40$\pm$0.48  & 69.02$\pm$0.84   && 41.32$\pm$1.31       & 41.19$\pm$1.29 \\ 
			\midrule
		\end{tabular}
	}
	$\dagger$~Results obtained by running the author's publicly available code 3 times with the same settings as the proposed method.\\
	$\ddag$~Includes the time required to train, which was done using information of the performance of 100 fully trained networks, which collectively required 4.16 training days to train.
\end{table*}

\FloatBarrier

\subsection{NAS-Bench-201}
NAS methods tend to be hard to reproduce, compare with other methods, and evaluate their real performance on common search spaces \cite{lindauer2020best}. Increases in search spaces size, result in increasing the number of possible networks that can be generated, which using performance estimation strategies that require some type of training makes exhaustively evaluating the performance of NAS methods extremely hard, ultimately resulting in evaluations using subsets of the whole search space (thousands of networks in a search space that can ultimately be unbounded). It is crucial that methods smooth the reproducibility process by adopting common training procedures and settings \cite{lindauer2020best}.

Recently, NAS benchmarks have been proposed, where the goal is to have a controlled setting, where information about the training and final performance of possible networks under the proposed search space is provided, allowing rapid prototyping and comparison between different NAS methods using the same search space, training procedures and hyper-parameters \cite{DBLP:conf/icml/YingKCR0H19, DBLP:conf/iclr/ZelaSH20, DBLP:journals/corr/abs-2008-09777,DBLP:conf/iclr/Dong020}.

In this work, we used NAS-Bench-201 \cite{DBLP:conf/iclr/Dong020} to evaluate the proposed method. NAS-Bench-201 provides information about trained networks in three different datasets: CIFAR-10, CIFAR-100 and ImageNet16-120, with fixed splits, and also provide results of several NAS methods under its constraints, allowing direct comparison. In this benchmark, the goal is to design cell-based architectures, where each cell is comprised of 6 edges and 4 nodes. All nodes receive an input edge from all the preceding nodes. The edges represent the possible operations, which are selected from a pool of 5 operations: (1) zeroize, which zeros the information, (2) skip connection, (3) $1\times1$ convolution, (4) $3\times3$ convolution, and (5) $3\times3$ average pooling layer. The number of possible operations and edges means that there are $5^6 = 15625$ possible cells. The final networks are comprised of a fixed macro skeleton, where a cell is a replicated block in the network, meaning that there are as many networks as possible cells, as the only change in the macro skeleton is the cell to be replicated.

\subsection{Results and Discussion}
\begin{figure}[H]
\vspace{-12pt}
  \centering
  \raisebox{-0.5\height}{\subfloat{\includegraphics[width=0.49\columnwidth, keepaspectratio]{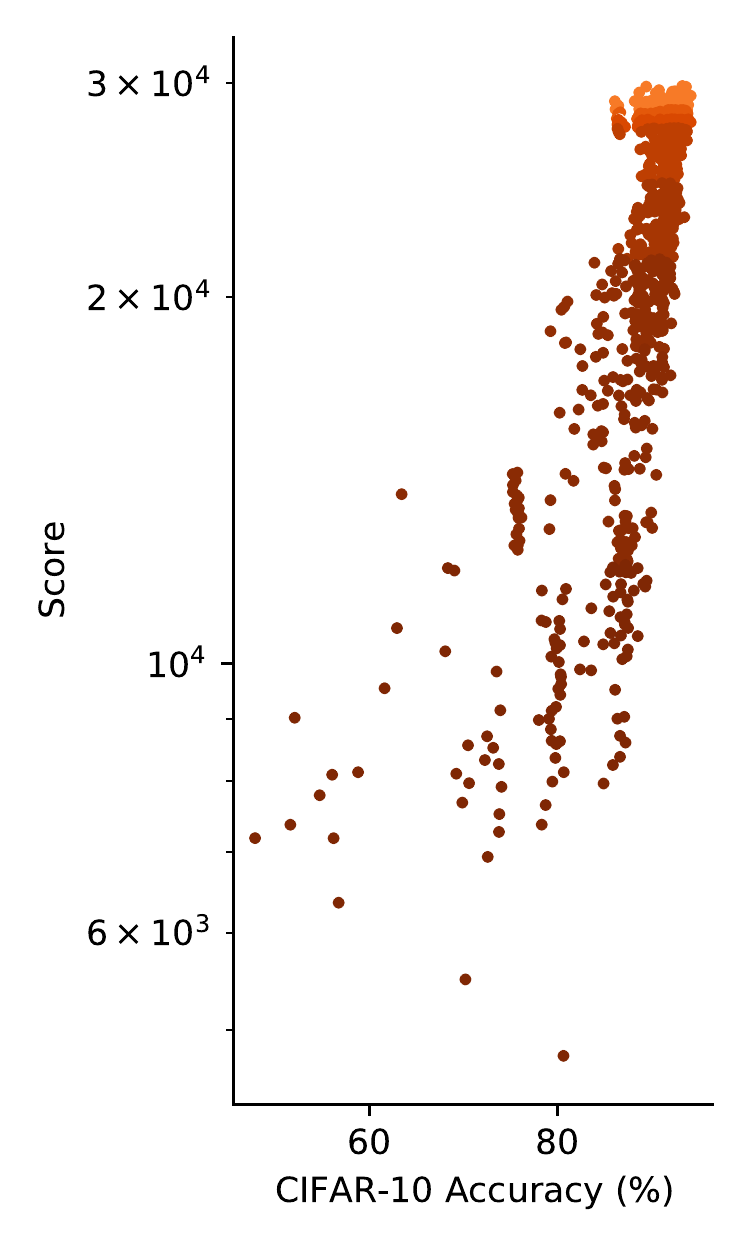}\label{fig:score_cifar10}}}
  \raisebox{-0.5\height}{\subfloat{\includegraphics[width=0.49\columnwidth, keepaspectratio]{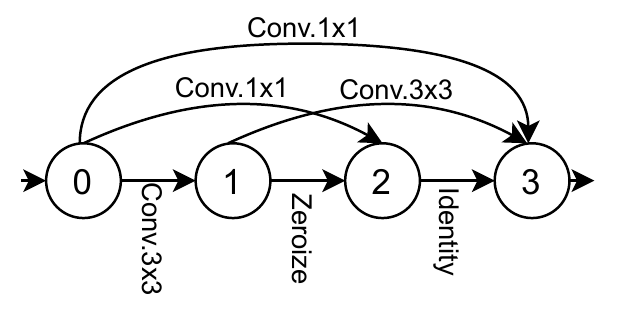}\label{fig:cell_cifar10}}}\\
  \vspace{-0.8em}
  \raisebox{-0.5\height}{\subfloat{\includegraphics[width=0.49\columnwidth]{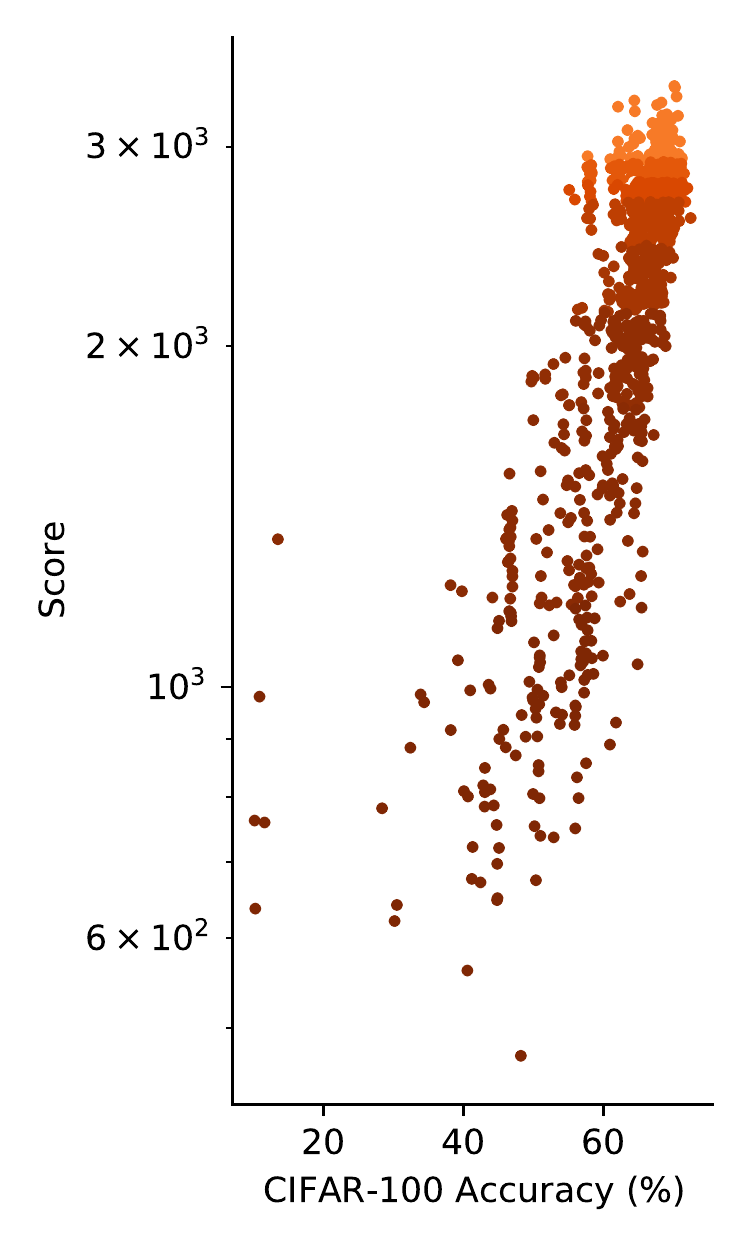}\label{fig:score_cifar100}}}
  \raisebox{-0.5\height}{\subfloat{\includegraphics[width=0.49\columnwidth]{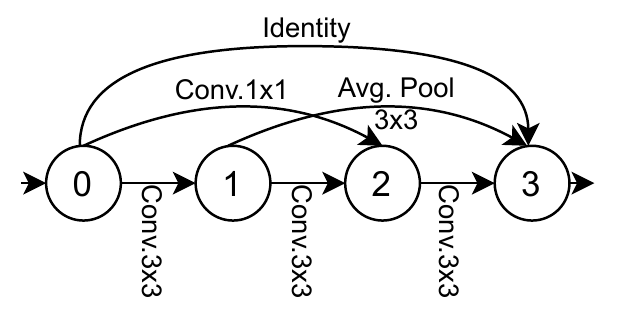}\label{fig:cell_cifar100}}}\\ 
  \vspace{-0.75em}
  \raisebox{-0.5\height}{\subfloat{\includegraphics[width=0.49\columnwidth]{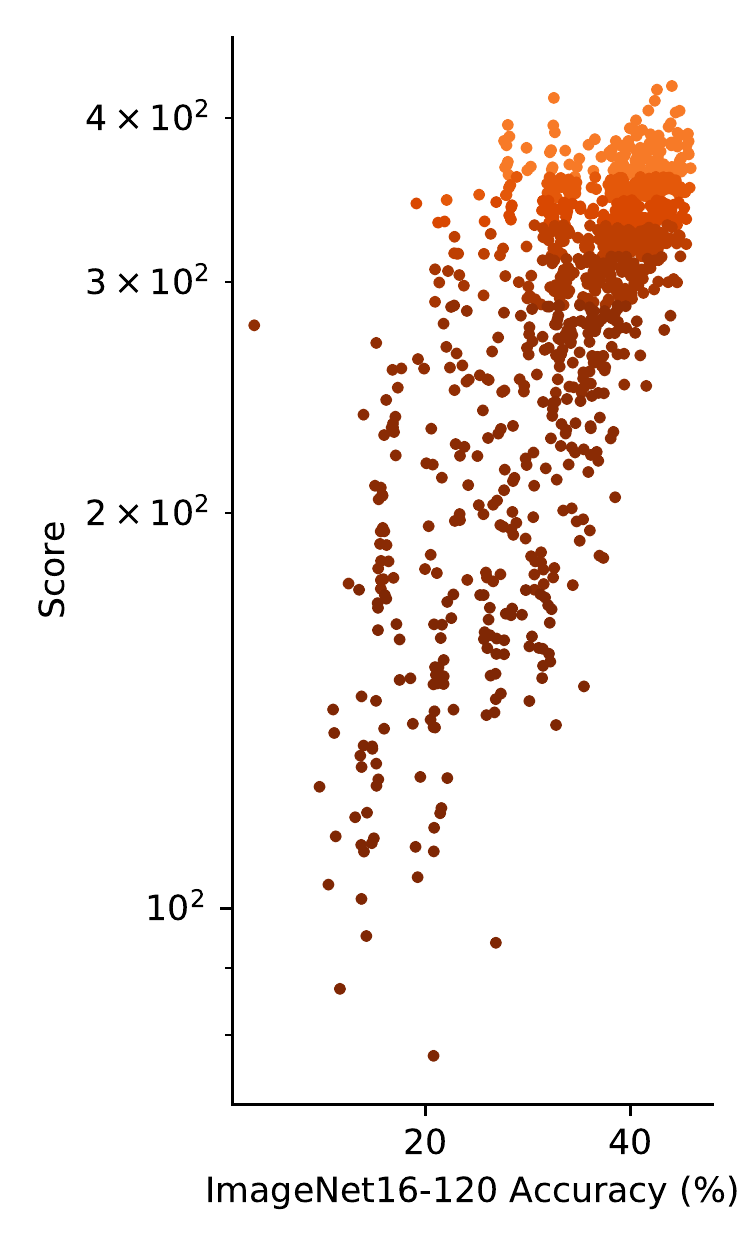}\label{fig:score_in16120}}}
  \raisebox{-0.5\height}{\subfloat{\includegraphics[width=0.49\columnwidth]{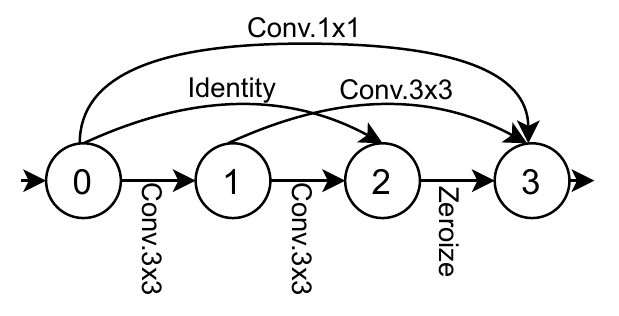}\label{fig:cell_in16120}}}\\

  \caption{Plots of scoring 1000 random untrained networks using the proposed method against the final accuracy when the networks are trained, using the 3 different datasets. On the right, the cell with the highest score, in each setting, is shown.\label{fig:plotscoredatasets}}

\end{figure}

% avaliação do método
First, we evaluate the effectiveness of the proposed method, by randomly sampling 1000 networks from each dataset of NAS-Bench-201 and score them to see the correlation between the score and the networks' performance when trained. This evaluation can be seen in Fig. \ref{fig:plotscoredatasets}, where the first row refers to CIFAR-10, the second to CIFAR-100 and the third to ImageNet16-120. On the left, it is possible to see a strong correlation between the score given by EPE-NAS and the network's accuracy once trained. This validates the proposed method by showing that scoring untrained networks highly is indicative of a higher performance than the networks scored lower. From this, it is also possible to see that by defining thresholds, e.g., $2\times10^4$ in CIFAR-10, we can efficiently weed out bad candidates, which is of utmost importance to methods based on evolution, where this method can be used to select the best networks from a pool of generated networks that will serve the purpose of generating the next iteration of the evolution. More, this information can also serve to guide the search of large search spaces, by directly indicating which network configurations are better. This is important because searching for networks in large search spaces, possibly unbounded, is extremely difficult and prone to converge to \textit{local minimas}, mainly due to lack of information about the search space which ultimately leads search methods to converge fast to the best networks initially sampled.

%juntar o método com random search, e comparar com outros métodos
Then, by combining EPE-NAS with a random search strategy, we can compare the effectiveness of a simple search strategy coupled with the proposed performance estimation strategy against other NAS methods. To perform this experiment, a network is randomly proposed, and instead of training it, we evaluate its performance by scoring the network. This setup requires no training, and we can perform this for different sample sizes ($N$, where $N$ represents the number of networks evaluated). Table \ref{table:benchmarking} shows the results for EPE-NAS with random search, and compares it with several methods. Methods that perform the search without weight sharing are shown in the first block, whereas weight sharing methods are shown in the second block. In the third block, we present the results for the proposed method and directly compare it with NAS-WOT \cite{mellor2020neural}, while also showing the optimal network in each setting where our method and NAS-WOT were evaluated. Finally, in the last block, we show a baseline method based on a Support Vector Machine (SVM). The SVM approach is indicative of a possible surrogate model that is trained with information of the 100 trained networks performances. The SVM input was created by computing a single correlation matrix of the batch and then calculating the eigenvalues of the matrix. Denote that for training the 100 networks, 4.16 days of GPU computation were required. After finalizing the SVM training, there is no need to further train any network, as the SVM infers the performance based on untrained networks' correlation matrix.

From this table, it is possible to see that our proposed method requires orders of magnitude less time to search for efficient networks, while both non-weight sharing and weight sharing incur in a large search time cost. Our method also achieves better results throughout all datasets than weight sharing methods, except for GDAS on CIFAR-10 and CIFAR-100. However, our method is more than $12500\times$ faster. The non-weight sharing methods outperform our method (random search coupled with the proposed performance estimation strategy), but our method is still on pair with them, being capable of achieving competitive results in all datasets. As for the direct comparison with NAS-WOT, in Table \ref{table:benchmarking} it is also shown that the proposed method outperforms NAS-WOT both in terms of inference, being faster in all settings, and in terms of accuracy, being capable of selecting high performant networks in CIFAR-10 and CIFAR-100 in the settings where the sample size is 10 and 100, and CIFAR-10, CIFAR-100 and ImageNet-16-120 for higher sample sizes (500 and 1000). It is important to note that for NAS-WOT, as sample size increases, it increasingly suffers from noise, increasing the gap between the chosen network accuracy and the optimal result and decreasing the performance compared to smaller sample sizes. The opposite happens with our method. As the sample size increases, our method is capable of selecting high performant networks without losing precision, which is of extreme importance, as it is improbable that optimal networks are present in small sample sizes. More focused on ImageNet-16-120, which is a dataset with more noise, due to the image sizes ($16\times16$) and the high number of classes, our method can select networks that attain excellent test accuracies, when compared with no weight sharing methods and NAS-WOT.

\begin{figure}[t]
    \centering
    \includegraphics[width=1\columnwidth]{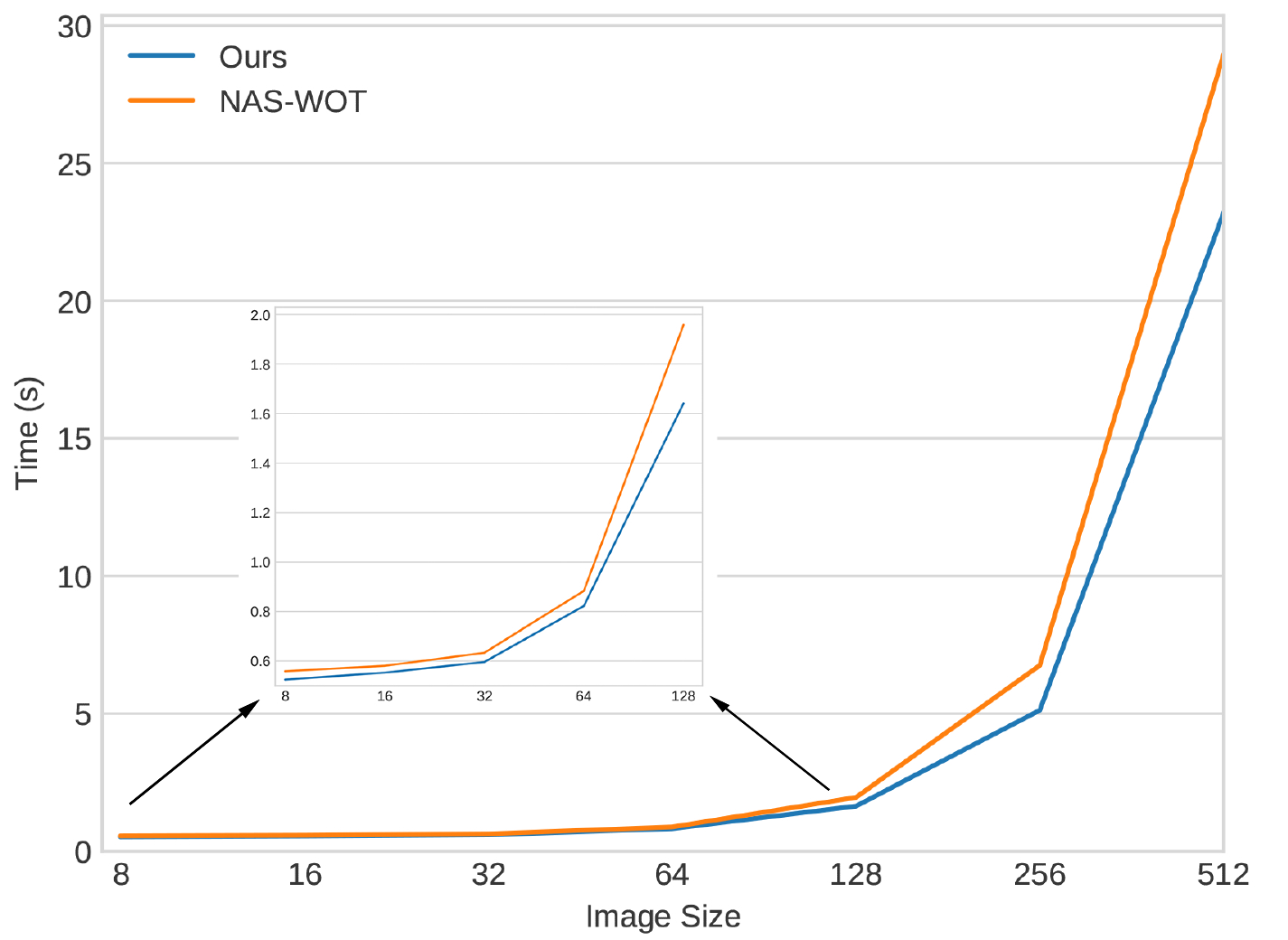}
    \caption{Comparison of the time, in seconds, required to score 1 network using our proposed method (in blue) against NAS-WOT, for different image sizes (x-axis). The image size represents the image's width and height, as the images evaluated are square and with 3 channels (RGB). \label{fig:plottimes}}
\end{figure}

As can be seen in the second column of Table \ref{table:benchmarking}, the execution time of our method is a great advantage, as it is capable of evaluating 1000 networks in 206 seconds. To further evaluate the gains in terms of execution time compared to NAS-WOT, we explored how both methods behave in scoring a network, with a batch size of increasingly different image sizes, which can be seen in Fig. \ref{fig:plottimes}. This evaluation shows that the proposed method consistently outperforms NAS-WOT, and that it is capable of evaluating images with sizes $256*256*3$ in approximately 5 seconds, meaning that the proposed method can also serve as an improvement for current NAS methods that solely search for networks in CIFAR-10, due to the reduced image size, and then transfer the best networks to ImageNet settings. Thus, NAS methods that were incapable of searching using larger datasets due to time complexity, can use EPE-NAS to search networks in larger datasets directly.

The reason why the proposed method is capable of outperforming NAS-WOT in terms of time is directly linked with the time complexity of creating a correlation matrix, which is highly dependant on the number of data points and features. By evaluating individual correlation matrices, one per class, we reduce each correlation matrix's size, allowing for faster computations.

Considering the mean time required to evaluate 1000 networks by our method (Table \ref{table:benchmarking}), EPE-NAS also allows exhaustive exploration of a search space, as the proposed method is capable of evaluating over 1 million architectures in just 2 days of GPU computing, under these settings. Therefore, this could be used to evaluate a search space's behaviour, giving information to the search method on how to start and proceed, which is a significant benefit when considering large, possibly unbounded, search spaces where information about their shape is limited.

% uma espécie de conclusão dos resultados obtidos
An important property of the proposed method is that it can easily be incorporated in almost any NAS method either as the sole method that evaluates networks or as a complementary method to perform mixed training, where the reward to update the controller parameters (Fig. \ref{fig:nasdiagram}) is a combination of complementary evaluations (e.g., EPE-NAS score combined with the inference/latency of the network in a mobile setting \cite{tan2019mnasnet, wu2019fbnet}). More, EPE-NAS is agnostic to the search method, as it focuses on the evaluation of networks, being the perfect addition to search methods that rely on information about generated networks or to guide the search, allowing the analysis of thousands of networks in seconds.

%%%%%%

\section{Conclusions}
\label{conclusions}
In this paper, we propose EPE-NAS, a performance estimation strategy that scores \textbf{untrained} networks with a high correlation to their trained performance. By leveraging information about the gradients of the output of a network with regards to its input, our method can accurately infer if the generated network is good in less than one second, being capable of evaluating thousands of networks in a matter of seconds. More, in this work, we have shown that using a simple random search coupled with the proposed estimation strategy, it is possible to sample high performant networks, in seconds, that can outperform many current NAS methods.

Our proposal can also contribute to allow NAS methods to search large search spaces, by providing an efficient way of extracting information about generated networks without requiring any training, and large databases, as our method is still very fast even in the presence of large image sizes. Furthermore, the proposed method is agnostic to the search strategy, allowing it to be integrated into almost any NAS method.

\section*{Acknowledgments}
This work was supported by `FCT - Fundação para a Ciência e Tecnologia' through the research grant `2020.04588.BD', partially supported by NOVA LINCS under grant `UID/EEA/50008/2019' and and partially supported by operation Centro-01-0145-FEDER-000019 - C4 - Centro de Competencias em Cloud Computing, cofinanced by the European Regional Development Fund (ERDF) through the Programa Operacional Regional do Centro (Centro 2020), in the scope of the Sistema de Apoio à Investigação Cientifíca e Tecnologica - Programas Integrados de IC\&DT.

\bibliographystyle{IEEEtran}
\bibliography{references}

\end{document}